\documentclass[sigconf]{acmart}
\usepackage{amsmath}
\usepackage{amsfonts}
\usepackage{algorithm}
\usepackage{algorithmic}
\usepackage{booktabs}
\usepackage{graphicx,xcolor}
\usepackage{hyperref}
\usepackage{float}
\usepackage{xspace}

\AtBeginDocument{%
  \providecommand\BibTeX{{%
    \normalfont B\kern-0.5em{\scshape i\kern-0.25em b}\kern-0.8em\TeX}}}

\copyrightyear{2022}
\acmYear{2022}
\setcopyright{acmcopyright}
\acmConference[MM '22] {Proceedings of the 30th ACM International Conference on Multimedia }{October 10--14, 2022}{Lisbon, Portugal.}
\acmBooktitle{Proceedings of the 30th ACM International Conference on Multimedia (MM '22), October 10--14, 2022, Lisbon, Portugal}
\acmPrice{15.00}
\acmISBN{978-1-4503-9203-7/22/10}
\acmDOI{10.1145/3503161.3548139}

\begin{document}

\begin{abstract}

Online social networks have stimulated communications over the Internet more than ever, making it possible for secret message transmission over such noisy channels. 
In this paper, we propose a Coverless Image Steganography Network, called \textbf{CIS-Net}, that synthesizes a high-quality image directly conditioned on the secret message to transfer. 
CIS-Net is composed of four modules, namely, the Generation, Adversarial, Extraction, and Noise Module. 
The receiver can extract the hidden message without any loss even the images have been distorted by JPEG compression attacks. 
To disguise the behaviour of steganography, we collected images in the context of profile photos and stickers and train our network accordingly. As such, the generated images are more inclined to escape from malicious detection and attack. 
The distinctions from previous image steganography methods are majorly the robustness and losslessness against diverse attacks.
Experiments over diverse public datasets have manifested the superior ability of anti-steganalysis.

\end{abstract}




\title{Image Generation Network for Covert Transmission in Online Social Network}
\author{Zhengxin You}
\email{zxyou20@fudan.edu.cn}
\affiliation{%
  \institution{Fudan University}
  \city{Shanghai}
  \country{China}
}

\author{Qichao Ying}
\email{shinydotcom@163.com}
\affiliation{%
  \institution{Fudan University}
  \city{Shanghai}
  \country{China}
}

\author{Sheng Li}
\authornote{Corresponding author.}
\email{lisheng@fudan.edu.cn}
\affiliation{%
  \institution{Fudan University}
  \city{Shanghai}
  \country{China}
}

\author{Zhenxing Qian}
\authornotemark[1]

\email{zxqian@fudan.edu.cn}
\affiliation{%
  \institution{Fudan University}
  \city{Shanghai}
  \country{China}
}

\author{Xinpeng Zhang}
\email{zhangxinpeng@fudan.edu.cn}
\affiliation{%
  \institution{Fudan University}
  \city{Shanghai}
  \country{China}
}

\maketitle

\section{Introduction}

Digital images have largely replaced the conventional photographs from all walks of life since the development of Online Social Networks (OSN) have make data sharing more convenient and efficient.
In the past decades, many efforts were made in mining the potential of data hiding within digital images. Such technology, named steganography, can benefit covert data transmission for military or medical applications or efficient cloud labeling. Steganalysis, as the adversary of steganography, is also widely studied to reveal the presence of steganography.

\begin{figure}[!t]
  \centering
  \includegraphics[width=0.45\textwidth]
  {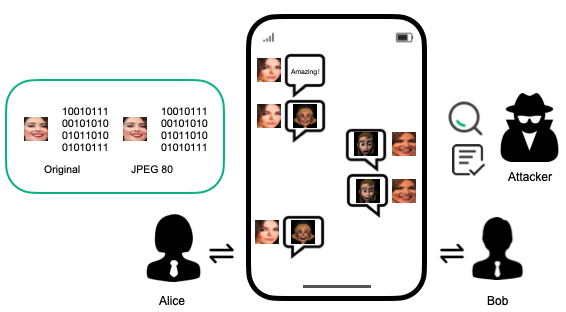}\\
  \caption{A practical application of our proposed scheme. Images generated by our method can be used as profile photos and stickers for covert transmission in online social network. After noise attack like JPEG Compression, secret message can still be extracted. Meanwhile, it is hard for attackers to detect our secret images.}
  \label{appfig}
\end{figure}
Traditional steganography methods ~\cite{filler2011minimizing} often need to choose a carrier, such as images that have abundant visual redundancy.
Generally, a data hider embeds secret messages into digital images and generates the corresponding container images that is close to the cover.
For example, content adaptive steganography~\cite{li2014new,holub2014universal,holub2012designing} heuristically design several kinds of embedding distortions to minimize the embedding loss. 
In recent years, deep networks are employed to boost the performance of image steganography~\cite{zhu2018hidden,ahmadi2020redmark,jing2021hinet,lu2021large}. For example, Zhu et al.~\cite{zhu2018hidden} and Ahmadi et al.~\cite{ahmadi2020redmark} developed robust watermarking scheme with strong robustness against a variety kinds of attacks. Jing et al.~\cite{jing2021hinet} and Lu et al.~\cite{lu2021large} are able to hide several images into a single host.
However, \cite{jing2021hinet,lu2021large} are fragile, and once the container images are attacked, the hidden information cannot be retrieved. It still remains difficult to simultaneously ensure a large capacity as well as robustness. Besides, these methods hide secret information based on modification towards the cover, and as a result, traces will inevitably be left during embedding and be detected by well-designed steganalyzer.

Compared with traditional image steganography methods, coverless image steganography methods do not make any modification to the carrier, which makes the previous steganalysis methods ineffective. 
With the progress of Generative 
Adversarial Networks (GAN)~\cite{karras2018progressive, brock2018large, karras2019style, zhang2019self}, realistic images can be automatically generated by neural networks and the quality of generated images is continuously improving. 
Therefore, scholars proposed several methods~\cite{hu2018novel, zhang2019generative, zhang2020generative} by using GANs for coverless image steganography. However, there still exists some weaknesses, among them the quality of generated images is still the heel of Achilles. 
% In this paper, we try to improve image quality with some tricks.
We observe that when people interact on social networks, they often use images such as profile photos and stickers. 
In ~\cite{souza2015dawn}, researchers find that users in OSN frequently post self-portraits on their profiles. In ~\cite{tang2019emoticon} and ~\cite{zhou2017goodbye}, researchers show that stickers become an integral part of most users’ activities.
Using these common images for steganography will further reduce the suspicion of attackers.
Therefore, we consider using GAN to generate images that can be used as profile photos and stickers in social networks for secret message transmission. We illustrate our steganography scenario in Fig.~\ref{appfig}. It should be noted that when images are uploaded on the social network, they may be attacked by JPEG Compression and some other noises. The noises added to the original images need to be taken into consideration in our pipeline.
To this end, we follow the idea of style-based generator~\cite{karras2019style} which can better disentangle secret message to synthesize natural images.

In this paper, we propose a Coverless Image Steganography Network (CIS-Net) for covert transmission on OSNs. 
In traditional covert transmission methods, the information is embedded by modifying the original image pixels. 
While in our method, we no longer use pixel modification, instead, we use different semantics to represent secret messages. 
Our network includes four modules: Generation, Adversarial, Extraction, and Noise Module. 
% We find that existing coverless image steganography methods generate images with low quality, and the generation task requires a lot of GPU resources. 
Through experiments we found that, the network tends to leave noise in the high-frequency area of generated images, resulting in poor visual effect. Therefore, for facial image datasets, we propose that surrounding areas of facial images can be cropped. Our experimental results show that the quality of images is higher when trained with cropped facial images.
In addition, we use the Facial Expression Research Group 2D Database (FERG-DB)~\cite{aneja2016modeling} to train our network, which is generated by MAYA software. The images in FERG-DB are very similar except for the expression, thus neural network converges quickly and generated images can be used as stickers in the social network. 
The image size used in our experiment is 32x32 and the embedding capacity is up to 32 bits per image. Although the image size used in our method is smaller than the previous robust coverless image steganography method, the embedding capacity(bit per pixel) and extraction accuracy of our method are higher. More importantly, because the image size we use is small, the network training needs less GPU memory, so we can retrain more models with less resources to resist attackers. 
The models retrained with different training datasets or network structure adjustments are difficult to be found by attackers. We will explain it in the experimental part. 
Compared with the image selection method, our model needs less storage space. Assuming that we want to transmit secret message of 32 bits, we need a database with 4294967296 images. However, our model only needs about 300MB of storage space, which greatly saves the cost.

The main contributions of our proposed method are as follows:

\begin{itemize}
    \item Compared with previous methods, CIS-Net can generate images with improved quality. Images generated by our method are close to natural images that reduces the suspicion of attackers.
	\item CIS-Net is robust against common noises in OSN such as JPEG Compression, where the recipient can extract original information with high accuracy.
	\item CIS-Net ensures high security for convert transmission, which can better evade the traditional steganalysis systems.
\end{itemize}

\section{Related Work}

\subsection{Image Steganography and Steganalysis}

Several effective image steganography were developed for the purpose of covert communication, copyright protection, etc.
For example, HUGO~\cite{holub2012designing} is a highly secured steganography system designed to minimize distortion to high-dimensional multivariate statistics. 
The Syndrome-Trellis Coding (STC)~\cite{filler2011minimizing} designed predefined embedding costs for all pixels or DCT coefficients. 
Volkhonskiy et al.~\cite{volkhonskiy2020steganographic}
first adopted deep networks to conduct data hiding within images. The performance is promising in both the authenticity of the generated images and the resistance to steganalysis systems. Recently, Jing et al.~\cite{jing2021hinet} and Lu et al.~\cite{lu2021large} proposed two similar works that employ Invertible Neural Networks (INN) to hide up to ten secret images into a single host image, therefore unveiling the potential of information hiding within natural images.

Steganalysis is to detect whether a targeted image contains secret data. For example, SPAM features~\cite{pevny2010steganalysis} uses Markov transition probabilities calculated by using adjacent pixels in eight directions to extract clues left by data hiding. 
SRNet~\cite{boroumand2018deep} designed a deep residual network to study the noise residual patterns inside stego images.
With the rapid development of image steganalysis as well as the fact that modification-based traditional image steganography will inevitably leave clues during data hiding, it is empirical to develop novel data hiding schemes that circumvent modifying the host.

\subsection{Coverless Image Steganography}

Traditional coverless image steganography methods are generally based on the selection of proper host image for covert data transmission. For example, Fridrich et al.~\cite{barni2011steganography} proposed an image selection method for information hiding. Image databases are established first and appropriate images are selected based on rules to represent secret message. Li et al.~\cite{li2018toward} proposed a method by using fingerprint synthesis to represent secret message. It is difficult for attackers to detect the existence of secret message from generated fingerprint images while embedding capacity is relatively small.

With the advancement of GAN technology, scholars proposed several methods based on GAN to generate images directly from secret message. Hu et al.~\cite{hu2018novel} proposed a generated adversarial network for image steganography, and secret message is mapped into a noise vector. To improve the quality of generated images, Yu et al.~\cite{yu2021improved} used attention module which also improves embedding capacity and extraction accuracy. Li et al.~\cite{li2021robust} uses a fixed Style-GAN to learn the mapping between secret message and generated images, but the extraction accuracy on the test set is only 50\%. Dong et al.~\cite{dong2021towards} considered the robustness and added a noise layer, but some speckle noises appear in the generated image. In \cite{cao2020coverless} and \cite{xue2021message}, attributes of generated images are used to represent the secret message. Cao et al.~\cite{cao2020coverless} used attributes of animation characters, such as hairstyles, hair colors, eye colors, and other features. Xue et al.~\cite{xue2021message} proposed multi-domain image translation to convert the original images to target domains such as smiling, narrow eyes, and black hair.

Though the results are impressive, the generated image quality needs to be improved. In addition, embedding capacity of robust coverless steganography is relatively small and real-world JPEG Compression attack is not taken into account.

\subsection{Robust Data Hiding with Deep Networks}

For a neural network, backward propagation works when each module is differentiable. However, some common image process operations such as JPEG Compression are non-differential. Zhu et al.~\cite{zhu2018hidden} proposed a method called HiDDeN which includes \textit{JPEG-Mask} and \textit{JPEG-Drop} to simulate real-world JPEG compression. 
However, this simulation is still quite different from the real-world counterpart in that the quantization table in real-world JPEG images can be customized and flexibly controlled by the quality factor as well as the image content. As a result, the neural networks can over-fit and lack real-world robustness. 
Jia et al.~\cite{jia2021mbrs} proposed mini-batch of real and simulated JPEG compression, which significantly improves the robustness. Liu et al.~\cite{liu2019novel} proposed a two-stage training strategy for non-differential operations. Encoder and decoder are trained separably to avoid the influence of JPEG Compression. In \cite{zhang2021towards}, Zhang et al. proposed a simpler and more effective method, which treats JPEG Compression noise as a constant tensor added to original images. In our method, we follow the pipelines in \cite{zhang2021towards} and \cite{liu2019novel} which work well in a real scenario.

\section{Methodology}

\subsection{Overview}

\begin{figure}[!t]
  \centering
  \includegraphics[width=0.5\textwidth]
  {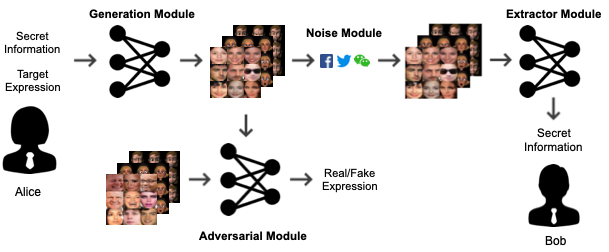}\\
  \caption{Framework overview. The Generation Module transforms the secret information into a secret image, which can be optionally controlled by the target expression. The Adversarial Module monitors the quality of the generated images by distinguishing them from natural images. The Noise module simulates image processing attacks in OSN and the Extractor Module gets the hidden message from the attacked image.}
  \label{methodfig}
\end{figure}

In this part, we will describe the proposed CIS-Net. 
Two models are used in our implementation and fig.~\ref{methodfig} explains our framework. One model is used to directly generate facial images, and the other model can control expressions of generated facial images. 
Our goal is to train a generation network that can map bits $B$ to a natural image $I$. JPEG Compression may be applied to the original image $I$ and a noised image $I_n$ is sent to the receiver. A trained information extractor can map image $I_n$ to original bits $B$. 
The core of our method is to map different secret messages to different semantic images. We use secret messages to control normalization of generator, different messages can be used to control different scales and biases of feature maps.

\subsection{Generation Module}

\noindent\textbf{Image preprocessing.} Through our experiments, we found that if we directly use the original facial image datasets like CelebA~\cite{liu2015faceattributes}, the generated images tend to have noise in texture-rich parts such as hair and the visual effect is not very good in the background part. That is to say, it is not easy for the network to learn high-frequency information. Details of image background and hair are more variable and complex. So we use a pre-trained face detection network MTCNN~\cite{zhang2016joint} to crop original facial images. Comparison between original images and cropped images is shown in Fig.~\ref{cropfig}.

\begin{figure}[!t]
  \centering
  \includegraphics[width=0.38\textwidth]
  {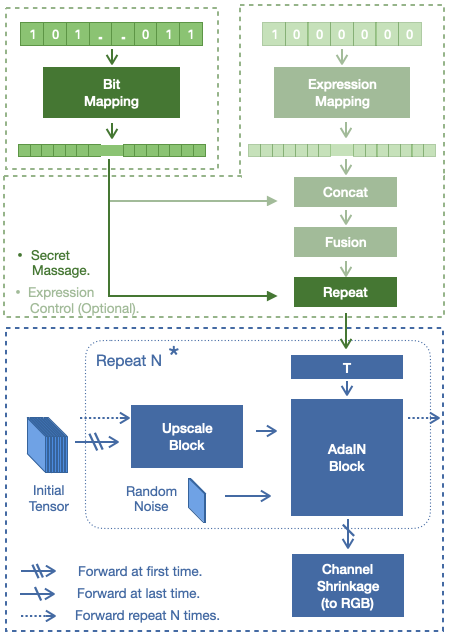}\\
  \caption{Generation Module. Secret messages are mapped into latent vectors which control adaptive instance normalization (AdaIN) to generate images. The expression mapping part is optional and is marked as gray in figure. Initial tensor is fed into network at first time and later the input of upscale block is replaced by output from AdaIN block. At last, output from AdaIN block is transformed to RGB images.}
  \label{generationfig}
\end{figure}

\noindent\textbf{Bit preprocessing.} We denote secret message as $B$, and we use $B$ to control different semantics of generated images. To this end, we map $B$ into a latent space that is suitable for semantics synthesis. As shown in Fig.~\ref{generationfig}, We map bits $B$ into latent vector $z_b$ through the bit mapping network. We directly expand $z_b$ repeatedly to obtain the same latent vectors, $z_1, z_2, ..., z_t$, which are respectively sent to different layers in the later semantic synthesis network.

For facial image generation with controllable expression, we use a one-hot vector $L'$ to control expressions. In this scenario, we also map the target domain label $L'$ to a latent space as shown in Fig.~\ref{generationfig}. After $L'$ is mapped into a latent vector $z_l$ through expression mapping network, we concatenate vector $z_b$ and $z_l$, and feed them into the later fusion network $F$ to obtain the final latent vector $z_f$. Similarly, we directly extend $z_f$ repeatedly to obtain $z_1, z_2, ..., z_t$.

\noindent\textbf{Semantic synthesis.} To generate robust images against attacks such as JPEG Compression, we consider synthesize secret messages from the perspective of semantics, because most semantic information remains unchanged after distortion. Following the adaptive instance normalization (AdaIN) proposed in \cite{karras2019style}, in our scheme, we use secret message to control scales and biases required in feature maps normalization as shown in Fig.~\ref{generationfig}. Image synthesis starts from a learned $4\times4\times512$ constant tensor, as shown in Fig.~\ref{generationfig}. After convolution operations, we upsample the original tensor to $8\times8$. Here, we can use upscale or ConvTranspose2d operation in PyTorch. We will talk about using different operations in the following security analysis. For $z_k$ obtained from mapping network where $k\in[1, 2, ... t]$, we map it to the space $Y$, $Y=(y_s, y_b)$, which has the same dimension as feature channels. The mapping network is realized through a fully connected layer $T_k$.

\begin{equation}
\begin{aligned}
ys, yb = Tk(zk).
\end{aligned}
\label{eq:eq3}
\end{equation}

Then, we use $y_s$ and $y_b$ for AdaIN to control different semantics. For each feature map $x_i$, the formula is as follows.

\begin{equation}
\begin{aligned}
AdaIN(x_i, y) = y_s \frac{x_i - \mu(x_i)}{\sigma(x_i)} + y_b,
\end{aligned}
\label{eq:eq4}
\end{equation}
where different $y_s$ and $y_b$ are used to control scales and biases of each feature map $x_i$. Since feature maps are normalized separately, images with different semantics can be generated according to different bits $B$. 

We add randomly generated noise $\emph{random}\_\emph{noise}$ in the process of image generation. After multiplying the channel-wise learnable weight $W$, noise is directly added to feature maps of each layer. The formula is in Eq.~(\ref{eq:eq5}). We find that the effect of random noise is obvious when the length of secret messages is 16. We will show it in the experimental part.

\begin{equation}
\begin{aligned}
x_i = x_i + w \times random\_noise.
\end{aligned}
\label{eq:eq5}
\end{equation}

The steps mentioned above are repeated $N$ times until the image expands to the size we want. Finally, we compress image channels and generate the RGB images we need.

\subsection{Adversarial Module}

Our adversarial module has two tasks: one is to distinguish original images from generated images, and the other is to distinguish the attributes of images. Therefore, our adversarial module has two outputs, which are denoted as the probability distributions $D_\emph{SRC}$ and $C_\emph{ATT}$ over sources and attribute labels respectively. Our discriminator $D$ and attribute classifier $C$ share the same feature extraction network. 

\noindent\textbf{Discriminator.} To make generated images look similar to original images, following WGAN~\cite{arjovsky2017wasserstein}, the loss function used for the discriminator and generator is shown in Eq.~(\ref{eq:eq6}).

\begin{equation}
\begin{aligned}
& L_\emph{advd} = -E[D_\emph{SRC}(I_{real})] + E[D_\emph{SRC}(G(B))], \\
& L_\emph{advg} = -E[D_\emph{SRC}(G(B))], \\
\end{aligned}
\label{eq:eq6}
\end{equation}
where original images are denoted as $I_\emph{real}$, and images synthesized by the Generation module $G$ under the control of bits $B$ is denoted as $G(B)$. We denote $D_\emph{SRC}(I_\emph{real})$ and $D_\emph{SRC}(G(B))$ as probability distribution over real and generated image sources. 
% The discriminator is trained to distinguish between real and fake images, while the generator optimizes the above objectives to make generated image distribution close to the real image distribution. 
In the training process, we use WGAN-GP~\cite{gulrajani2017improved} for optimization. Similarly, for the generator which can control attributes, we only need to replace $G(B)$ with $G(B, L')$ in Eq.~(\ref{eq:eq6}). $G(B, L')$ represents images generated under the control of attribute tag $L'$ and bits $B$.

\noindent\textbf{Attribute classifier.} The attribute classifier is optional and can be used to control the generated image with specific attributes when needed. We denote $C_\emph{ATT}(L| I_\emph{real})$ as probability distribution over attribute labels $L$. For the training of the adversarial module, the loss function is shown in Eq.~(\ref{eq:eq7}). This loss enables the attribute classifier to learn from real images.

\begin{equation}
\begin{aligned}
& L_\emph{attd} = E_{I_\emph{real}, L}[-log(C_\emph{ATT}(L | I_\emph{real}))]. \\
\end{aligned}
\label{eq:eq7}
\end{equation}

For the Generation module $G$, the attribute classification loss is shown in Eq.(~\ref{eq:eq8}), where we want Generator $G$ to synthesize images with corresponding attribute tags $L'$. The following objective is minimized when $G$ is training, so that generated images can be classified correctly.

\begin{equation}
\begin{aligned}
& L_\emph{attg} =  E_{B, L'}[-log(C_\emph{ATT}(L' | G(B, L')))].  \\
\end{aligned}
\label{eq:eq8}
\end{equation}

We denote $G(B, L')$ as images generated under the control of bits $B$ and attribute tag $L'$.

\subsection{Noise Module}

The noise module is used to improve the robustness of generated images. In previous work, scholars proposed several methods to solve the backward propagation problem of non-differential operations. In this paper, we follow the pipeline proposed in ~\cite{zhang2021towards} which views noise caused by JPEG Compression as constant tensor. We first truncate the gradient of original images $I$ to get $I_\emph{ori}$, then we apply JPEG compression on $I_\emph{ori}$ with different quality factors to get noised image $I_\emph{JPEG}$. Residuals between original images and noised images can be calculated with $\emph{diff}=I_\emph{ori}-I_\emph{JPEG}$. Noises brought by JPEG Compression can be viewed as constant $\emph{diff}$ added to original images $I$. Therefore, we can simulate JPEG Compression as well as other non-differential operations. In this module, the Identity layer and JPEG Compression with quality factors $90$, $80$, $70$, $60$, and $50$ are used for training. 

\subsection{Extractor Module}

In order to extract original bit information from image semantics, we introduce an information extractor E, which is composed of a series of convolution layers and fully connected layers. L2 norm loss is used as our extractor loss.

\begin{equation}
\begin{aligned}
& L_{e} =  E_{B}[||B - E(N(G(B)))||_{2}],  \\
\end{aligned}
\label{eq:eq9}
\end{equation}
where $N(G(B))$ denotes the image attacked by the noise layer. With this loss, $E$ learns to extract the original bit information and $G$ can generate images with semantics that can resist noise attacks. Similarly, for G which can control attributes, we replace $G(B)$ with $G(B, L')$ in the above formula.

\subsection{Objective Function and Training Details}

The above-mentioned Generator, Adversarial, Extraction, and Noise module constitute our coverless image steganography network. The overall loss function is listed in Eq.~(\ref{eq:eq10}), and the network is trained in an adversarial manner.

\begin{equation}
\begin{aligned}
& L_{d} =  \lambda_{1} * L_\emph{advd} + \lambda_{2} * L_\emph{attd},  \\
& L_{g} =  \lambda_{3} * L_\emph{advg} + \lambda_{4} * L_\emph{attg} + \lambda_{5} * L_{e}.  \\
\end{aligned}
\label{eq:eq10}
\end{equation}

We generate small facial images with the size of $32\times32$, and embedding capacity is set as $16$ bits to $32$ bits per image. Since generated images are small, networks can be trained with less GPU memory. Our models are divided into two types: one is to directly generate facial images, and the other is to generate images with desired expressions. Specifically, we use a 7-bit one-hot vector to control seven kinds of expressions(Surprise, Sadness, Neutral, Joy, Anger, Fear, and Disgust). Our implementation is based on Pytorch and uses NVIDIA RTX 3090. For the model directly generating facial images, we initially set hyperparameters $\lambda_1$, $\lambda_3$, $\lambda_5$ as $1$, $1$, $10$, and The penalty parameter of WGAN-GP is set as 50. For the model which controls expression, the hyperparameters $\lambda_1$, $\lambda_2$, $\lambda_3$, $\lambda_4$, $\lambda_5$, are set as $1, 1, 2, 0.1, 10$ initially. When $L_{attg}$ is smaller than 0.4, we set $\lambda_4$ as 0. Adam optimizer is used to optimize our model parameters and batch size is set to be 32. In the beginning, the learning rate is set as $1e-4$, and other hyper-parameters are set as default. In the noise module, we use JPEG Compression operations provided in the Python PIL package.

\section{Experiments and Discussions}

\subsection{Experimental Settings}
\begin{figure}[!t]
  \centering
  \includegraphics[width=0.38\textwidth]
  {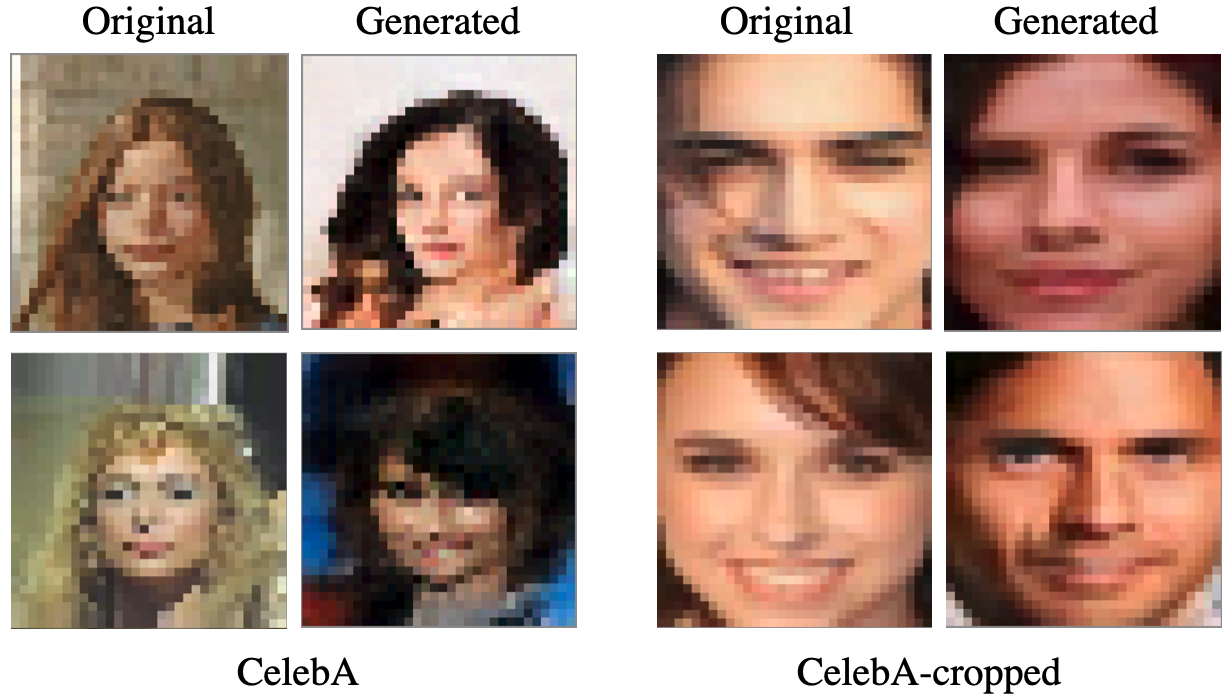}\\
  \caption{Comparison of generated images with/without image preprocessing. Noises can be found in textured areas if we directly train the network without image preprocessing.}
  \label{cropfig}
\end{figure}

We use CelebA~\cite{liu2015faceattributes}, LFW~\cite{huang2008labeled}, FFHQ~\cite{karras2019style}, FERG-DB~\cite{aneja2016modeling} as the training datasets. 
% CelebA has $202599$ images of $10177$ different people. LFW has $13233$ images of $5749$ different people, and FFHQ has $70000$ images of higher quality. 
For original images, we downsample them to the size of $32\times32$ and original images have some noise in texture-rich parts. It should be noted that network for small images generation tends to leave specific noise in the high-frequency section as shown in the above-mentioned Fig.~\ref{cropfig}. Therefore, We use a pre-trained MTCNN~\cite{zhang2016joint} to get cropped facial images. After removal of the external area of original images, we can use the cropped images to generate an image of higher quality. We denote cropped image datasets as CelebA-cropped, LFW-cropped, FFHQ-cropped. FERG-DB dataset has a total of $55767$ annotated face images. This dataset includes six cartoon characters(Aia, Mery, Bonnie, Ray, Jules, and Malcolm). Each cartoon character has seven kinds of expressions(Surprise, Sadness, Neutral, Joy, Anger, Fear, and Disgust). In our implementation, we use images of each character to train networks separately.

\noindent\textbf{Evaluation Metrics.}
For visual quality evaluation, we use Fréchet inception distances(FID)~\cite{heusel2017gans} as the evaluation metric, which can measure the distance between real and generated images.

\noindent\textbf{Benchmark.}
We employ state-of-the-art coverless steganography method for visual quality comparison~\cite{yu2021improved}. In addition, We compare our method with previous robust coverless steganography methods~\cite{cao2020coverless, xue2021message} to validate performance of robustness.

\begin{figure}[!t]
  \centering
  \includegraphics[width=0.48\textwidth]
  {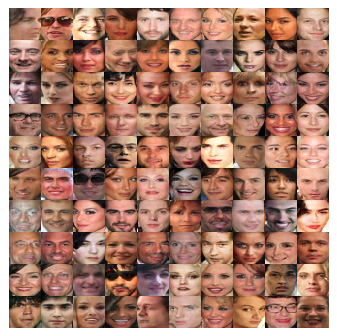}\\
  \caption{Examples of generated facial images. The length of the secret message is set as 32-bit.  }
  \label{celebafig}
\end{figure}

\begin{figure*}[!t]
  \centering
  \includegraphics[width=0.85\textwidth]
  {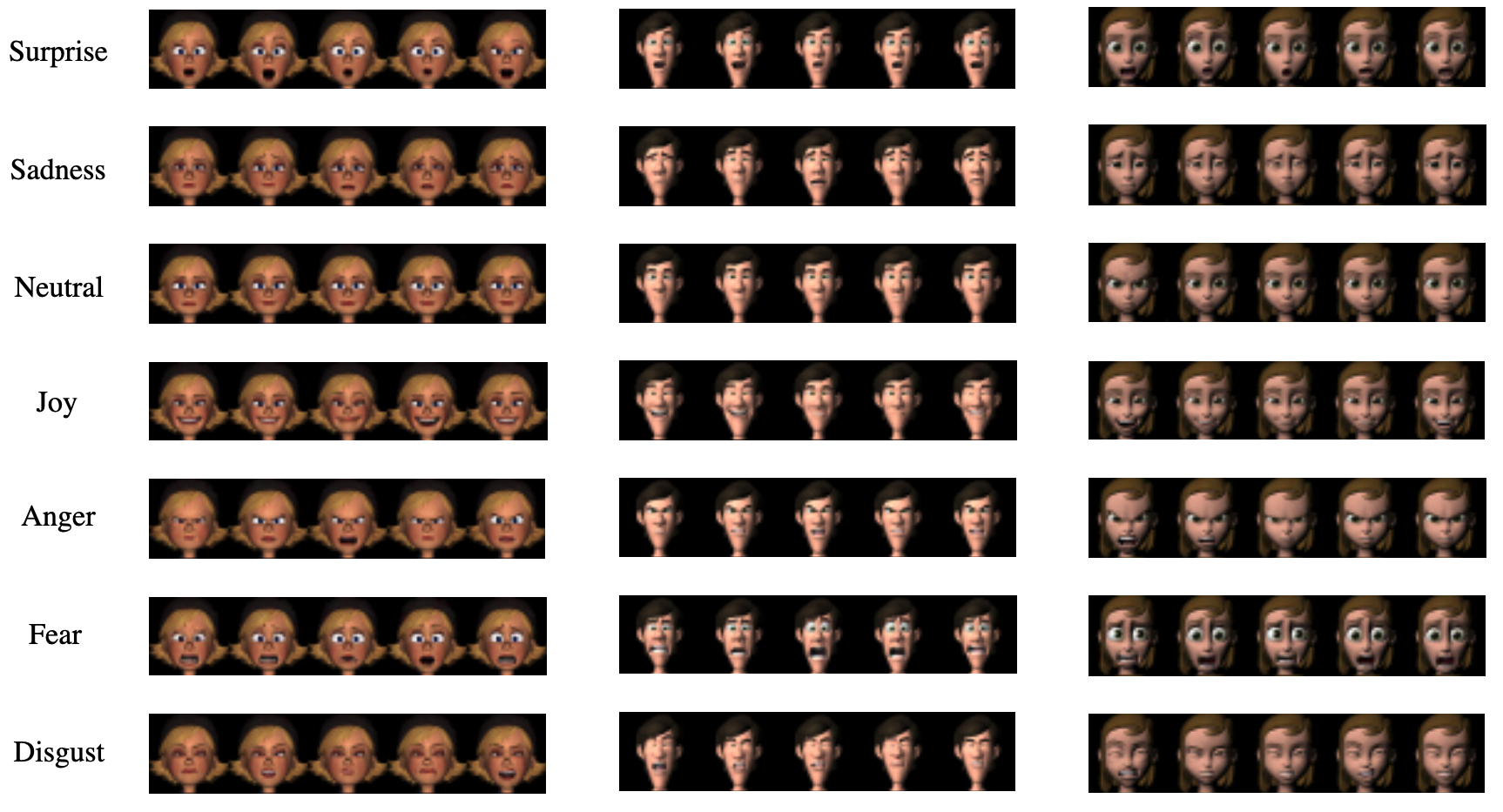}\\
  \caption{Example of generated images which can be used as stickers. From left to right are images synthesized using Bonnie, Malcolm, and Aia datasets. The secret message is set as 20-bit. Expression tags (Surprise, Sadness, Neutral, Joy, Anger, Fear, and Disgust) are used to control the generation of images.}
  \label{expressionfig}
\end{figure*}

\begin{table}[t]
    \centering
    \caption{Visual quality evaluation based on FID. Classiﬁcation accuracy is used to evaluate the accuracy of expression generation.}
    \begin{tabular}{lcc}
        \toprule
        Model & FID & Classiﬁcation Accuracy \\
        \midrule
        CelebA-cropped-32 & 6.20 & - \\
        CelebA-16 & 6.94 & -\\
        Bonnie-20 & 8.24 & 0.9620\\
        Ray-20 & 5.57  & 0.9763\\
        Mery-20 & 7.03  & 0.9497 \\
        Jules-20 & 5.28  & 0.9612 \\
        Aia-20 & 9.05  & 0.9234\\
        Malcolm-20 & 4.57  & 0.9782\\
        Yu et al.~\cite{yu2021improved} & 30.81 & -\\
        \bottomrule
    \end{tabular}
    
    \label{qualitytable}
\end{table}

\subsection{Comparisons}

\begin{table*}[t]
    \centering
    \caption{Results of robustness against JPEG Compression. }
    \begin{tabular}{lccccccccccc}
        \toprule
        Model & Original & JPEG90 & JPEG80 & JPEG70 & JPEG60 & JPEG50 & JPEG95 & JPEG85 & JPEG75 & JPEG65 & JPEG55\\
        \midrule
        CelebA-cropped-32 & 0.9798 & 0.9781 & 0.9752 & 0.9701 & 0.9638 & 0.9546&  0.9792 &0.9770 & 0.9728 & 0.9672 & 0.9594\\
        CelebA-16 & 0.9993 & 0.9986 & 0.9979 & 0.9967 & 0.9951 &   0.9913 & 0.9990 &  0.9983 &  0.9972 &  0.99575 & 0.9931\\
        Ray-20 & 0.9901 & 0.9911 & 0.9861 & 0.9771 & 0.9667 & 0.9540 & 0.9910 & 0.9883 & 0.9786 & 0.9684 & 0.9541\\
        Mery-20 & 0.9873 & 0.9843 & 0.9745 & 0.9624 & 0.9487 & 0.9263 & 0.9807 & 0.9784 & 0.9639 & 0.9546 & 0.9351\\
        Aia-20 & 0.9768 & 0.9811 & 0.9749 & 0.9667 & 0.9547 & 0.9336 & 0.9790 & 0.9771 & 0.9676 & 0.9569 & 0.9409 \\
        Bonnie-20 & 0.9747 & 0.9798 & 0.9741 & 0.9639 & 0.9517 & 0.9421 & 0.9777 & 0.9747 & 0.9666 & 0.9552 & 0.9443 \\
        Jules-20 & 0.9575 & 0.9685 & 0.9634 & 0.9580 & 0.9545 & 0.9220 & 0.9630 & 0.9596 & 0.9562 & 0.9422 & 0.9292 \\
        Malcolm-20 & 0.9706 &  0.9672 &    0.9564 &    0.9436 &    0.9239 &    0.9031 &   0.9637 &   0.9578 &   0.9391 &    0.9262 &   0.9014 \\
    
        \bottomrule
    \end{tabular}
    
    \label{jpegtable}
\end{table*}

% $50^\circ$ 
\begin{table*}[t]
    \centering
    \caption{Results of robustness against other noises and capacity.}
    \begin{tabular}{lccccccccccc}
        \toprule
        Model & Capacity(bpp) &  Rotation & Gaussian Noise & Salt \& Pepper & Speckle & Median Filter & Mean Filter & Gaussian Filter \\
        \midrule
        CelebA-16 & $1.5\times10^{-2}$ & 0.9818 & 0.9674 & 0.9893 & 0.9892 & 0.9815 & 0.9812 &  0.9919  \\
        Xue et al.~\cite{xue2021message} & $6.71\times10^{-4}$  & 0.7818 & 0.9157 & 0.9220 & 0.9220 & 0.9236 & 0.9184 & 0.9216\\
        Cao et al.~\cite{cao2020coverless} & $8.54\times10^{-4}$ & 0.8324 & 0.0362 & 0.5096 & 0.0893 & 0.8187 & 0.7576 & 0.8257\\
        \bottomrule
    \end{tabular}
    \label{noisetable}
\end{table*}

\noindent\textbf{Visual quality. }
For each model, we generate the same number of images in training datasets to calculate FID. Models with a 32-bit secret message as input and using CelebA as a training dataset are denoted as CelebA-32. Similarly, models trained with 16-bit secret message and LFW-cropped training dataset are denoted as LFW-cropped-16. We can find that CelebA-16 images have more noise than CelebA-cropped-32 images as shown in Fig.~\ref{cropfig}. More examples from CelebA-cropped-32 are shown in Fig.~\ref{celebafig}.  Compared with previous methods, images generated by our method are more natural and FID is shown in Table~\ref{qualitytable}. The experiment shows that our visual quality is better.

For models which can control expressions, we use every single character of FERG-DB to train the network respectively. Models trained with 20-bit secret message and using Bonnie as a training dataset are denoted as Bonnie-20. The experimental results are shown in Fig.~\ref{expressionfig} and FID results are shown in Table~\ref{qualitytable}. To evaluate whether our model can control facial expressions according to the input vector, we calculate the classiﬁcation accuracy of generated images, and the results are shown in the Table~\ref{qualitytable}. Experiment shows that our model achieves good performance.

\noindent\textbf{Embedding Capacity. }
Our embedding capacity is between 16 bits and $32$ bits per image. Although the embedding capacity of a single image is small, the metric bit per pixel (bpp) is high. We can splice multiple $32\times32$ images, then we can transmit a large amount of information as proposed in \cite{cao2020coverless}. In addition, since our generated images can be sent as stickers which are frequently used in social network chatting, we can convey more information through multiple transmissions. Compared to the previous robust coverless steganography methods, our embedding capacity is larger and the results are shown in Table~\ref{noisetable}.

\noindent\textbf{Robustness against JPEG Compression. }
In this section, we show experimental results of information extraction accuracy. To the best of our knowledge, we are the first to consider real JPEG Compression in coverless image steganography based on GANs. During training, the quality factors of JPEG Compression we used are $90, 80, ..., 50$, and quality factors are set as $90, 85, 80, ..., 55$ during testing. The extraction accuracy is shown in Table~\ref{jpegtable}. We can find that after applying JPEG Compression of different quality factors on original images, we can still extract secret message from the noised image. Since we can not achieve 100\% extraction accuracy, error correction codes can be used in real-world applications. 

\noindent\textbf{Robustness against Other Noises. }
For typical noises other than JPEG compression, we test the proposed scheme with image rotating , addition of Gaussian noise, addition of Salt $\&$ Pepper noise, addition of Speckle Noise, Gaussian blurring, Mean filtering and Median filtering, which was employed in \cite{cao2020coverless} and \cite{xue2021message}. We use model CelebA-16 to compare and follow \cite{liu2019novel} to enhance our robustness. The network is trained under JPEG Compression attack without the above-mentioned noises. Therefore, to resist these new attacks, we only need to train a new extractor separately while the generator is fixed. As shown in Table~\ref{noisetable}, our model not only extracts bits with higher accuracy but also embeds more secret message. Meanwhile, it demonstrates that our model has good generalization ability. When there are new channel attacks, we can directly train a new extractor to resist. Since our image size is relatively small and large filters will make images lose too much semantic information, we do not use $5\times5$ and $7\times7$ filters which are used in \cite{xue2021message}.

\begin{figure*}[!t]
  \centering
  \includegraphics[width=0.9\textwidth]
  {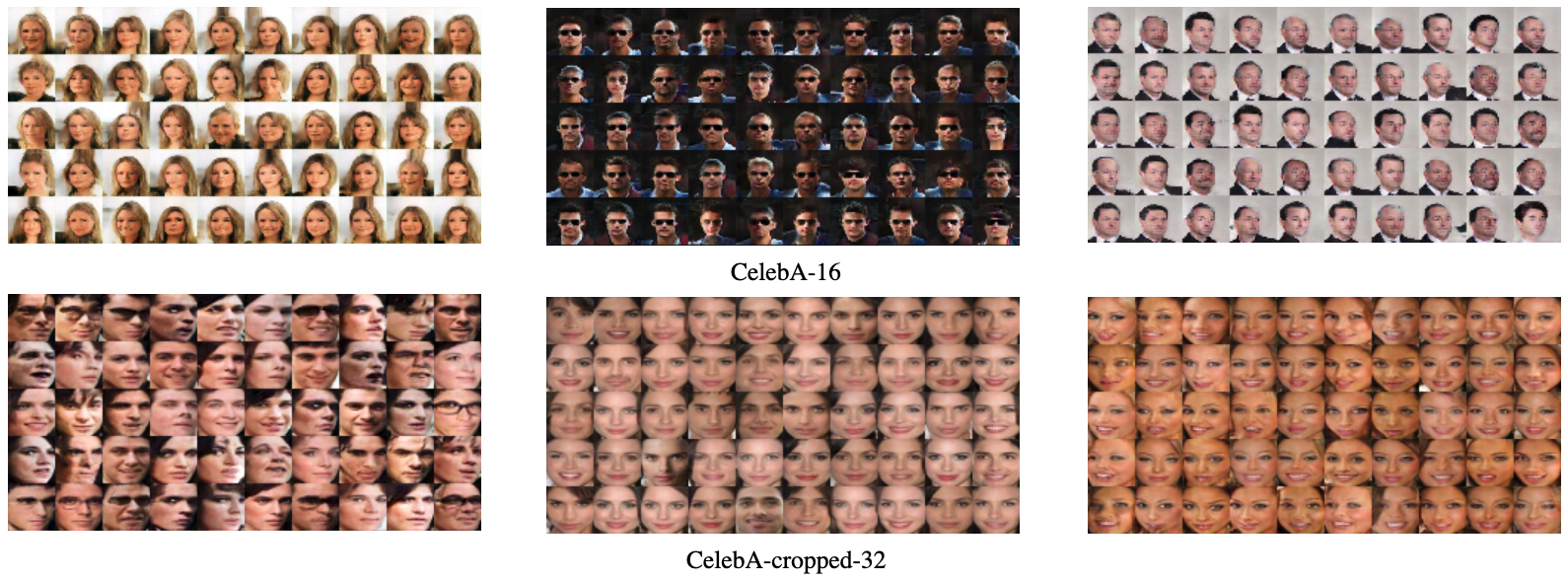}\\
  \caption{Images generated by the same secret message from CelebA-16 are shown in the first row. Images generated by secret messages with same first 26 bits from CelebA-crop-32 are shown in the second row.}
  \label{discussionfig}
\end{figure*}

\subsection{Security Analysis}

For GAN image detection, scholars proposed several methods~\cite{wang2020cnn, yu2019attributing} to distinguish real and fake images, but there are a large number of GAN images in online social networks. Obviously, it is not appropriate to directly detect whether images are real or fake in a steganography scenario. What we need to analyze is whether image dataset A generated by a normal GAN can be distinguished from image dataset B generated from secret message. Yu et al.~\cite{yu2019attributing} found that the images generated by different GANs can be distinguished. As long as network structures, training datasets, or random seeds of models are different, the generated images can be distinguished. In our experiment, we find that images from A and B can be classified and the results verify that Yu et al. reported. So what we need to test is the generalization performance of the detection network. We consider this scenario: Attackers trains a detection model  $\emph{DM}$ based on dataset $A$ and $B$. Whether image datasets $C, D, E...$ generated by other steganography models can be detected by $\emph{DM}$. We use two types of detection models for analysis, one is a simple convolution network $\emph{DM}_C$, and the other is a modified network $\emph{DM}_S$ from SRNet~\cite{boroumand2018deep} which is often used for steganalysis to fit our image size, i.e., $32\times32$.

We denote the model with input sampled from Normal distribution as M1, the model using secret message to generate images as $M_2$. We use images $I_1$, $I_2$ generated by $M_1$, $M_2$ to train detection model ${DM}_S$, ${DM}_C$. For the test dataset, we use images generated from models with a slight change, such as different training datasets and different network structures. Specifically, we use the following models.

\begin{itemize}
\item $M_3$: Retrained model with no other changes.
\item $M_4$: Deconvolution layer is used to replace upscale in the last upsample step.
\item $M_5$: Number of convolution kernel is reduced.
\item $M_6$: FFHQ-cropped is used as training dataset.
\item $M_7$: LFW-cropped is used as training dataset.
\end{itemize}

We denote images generated from $M_3$, ..., $M_7$ as $I_3$, ..., $I_7$, respectively. DM trained with $I_1$, $I_2$ is used to judge $I_3$, ..., $I_7$. The detection accuracy results are shown in Table~\ref{secure1table}. All the results are calculate ten times and average value is taken.

\begin{table}[t]
    \centering
    \caption{Security Analysis for images generated for profile photos.}
    \begin{tabular}{lccccccc}
        \toprule
        Model $M1$ & $M2$ & $M_3$ & $M_4$ & $M_5$ & $M_6$ & $M_7$ \\
        \midrule
         $\emph{DM}_C$ 1.00 & 1.00 & 0.52 & 0.52 &  0.57  &  0.49 &  0.37  \\
        % Xception & \\
         $\emph{DM}_S$  1.00 & 1.00 &  0.63 & 0.48  &  0.63 &   0.36 &  0.42 \\
        \bottomrule
    \end{tabular}
    \label{secure1table}
\end{table}

\begin{table}[t]
    \centering
    \caption{Security Analysis for images generated for stickers.}
    \begin{tabular}{lcccccccc}
        \toprule
        Model $M_{1}'$ & $M_{2}'$ & $M_{3}'$ & $M_{4}'$ & $M_{5}'$ & $M_{6}'$ & $M_{7}'$ & $M_{8}'$ \\
        \midrule
        $\emph{DM}_C$ 1.00 & 1.00 &  0.47 &   0.42  & 0.22  & 0.01 & 0.56 &  0.49 \\
        % Xception & 0.64  & 0.41 &  0.39 &  0.78 & 0.64  &  0.79 \\
        $\emph{DM}_S$ 1.00 & 1.00 & 0.20  & 0.11  &  0.10 &  0.23  &  0.26 & 0.39 \\
        \bottomrule
    \end{tabular}
    \label{secure2table}
\end{table}

Similarly, for generation network which can control expression, we also use the above method to test. Models are denoted as follows:
\begin{itemize}
\item $M_{1}'$: GAN with input sampled from Normal distribution.
\item $M_{2}'$: Bonnie dataset is used as training dataset.
\item $M_{3}'$: Secret message length is set to be different.
\item $M_{4}'$: Aia dataset is used as training dataset.
\item $M_{5}'$: Jule dataset is used as training dataset.
\item $M_{6}'$: Mery dataset is used as training dataset.
\item $M_{7}'$: Malcolm dataset is used as training dataset.
\item $M_{8}'$: Ray dataset is used as training dataset.
\end{itemize}

\begin{figure}[!t]
  \centering
  \includegraphics[width=0.42\textwidth]
  {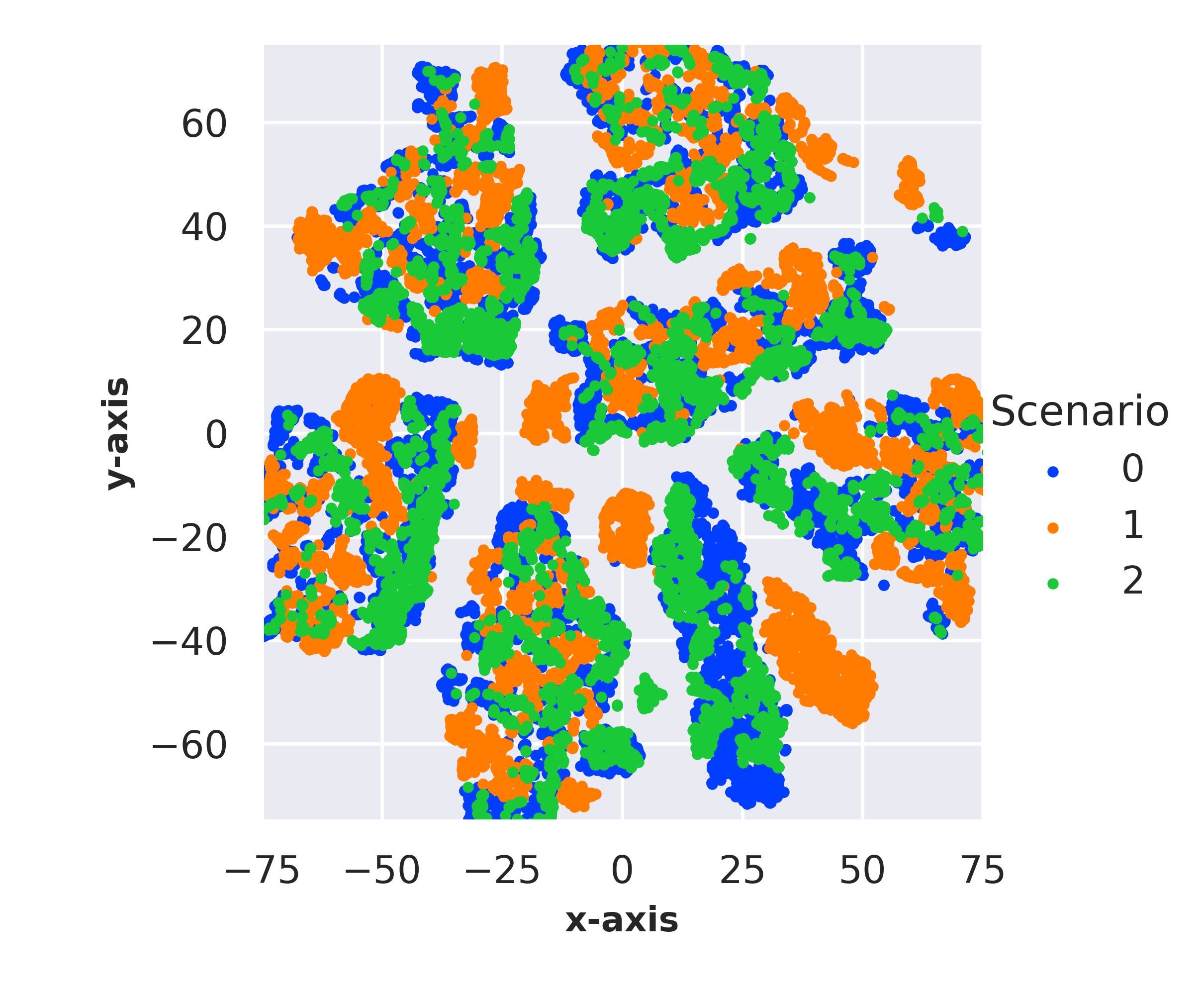}\\
  \caption{t-SNE visualization for an explanation. Where $0$ and $1$ represent images generated by two different Malcolm-16 models, $2$ represents images generated by normal GAN trained with Malcolm dataset. }
  \label{tsnefig}
\end{figure}

$I_{1}'$, ..., $I_{8}'$ are denoted as images generated from $M_{1}'$, ..., $M_{8}'$, respectively. The testing results are shown in Table~\ref{secure2table}. We find that the accuracy in test set sometimes reached 100\%, and sometimes reached 0\%. We think it is because $DM$ learns the distribution of datasets. If the distribution of the test image set is closer to $I_{1}'$, it is easy to be judged as image without secret message. Otherwise it is detected as image with secret message. We use t-SNE~\cite{van2008visualizing} to map images to two-dimension space for explanation. Images are firstly processed by three SRM filters~\cite{fridrich2012rich}. As shown in Fig.~\ref{tsnefig}, $0$ and $1$ are images with secret message while $2$ represents images from normal GAN. We find that images from $0$ that are with secret message are closer to $2$.

\section{Conclusion and Discussion}

In this paper, we propose a method for convert transmission in online social network.
Experimental results show that high-quality images can be generated from our methods and extraction accuracy is guaranteed after noise attack. The flexibility of model training makes it difficult for attackers to detect our secret message. 
Although we have considered most common noises, new unknown noises may reduce accuracy. We will try to study it in future work.

To figure out how our coverless image steganography method works, we test our model with same secret message as input. In previous work~\cite{cao2020coverless, xue2021message}, scholars use notable features to transmit secret message. Since our model also uses secret message to control semantics, we investigate implied features space our model generated. As shown in Fig.~\ref{discussionfig}, we use images from CelebA-16 and CelebA-cropped-32 for illustration. For CelebA-16, we set secret message to be the same in the first row. We find that facial images generated with the same secret message share common features in the background, skin color, posture, and so on. The only difference between these images is the different noise added in the Generation module. For 16-bit secret message, the network needs to learn $2^{16}$, that is 65536 different features. While for 32-bit secret message,  we find that in this case noise added in the Generation module almost does not work and the network tends to directly generate $2^{32}$ different facial images. In the second row, we set secret messages with same first 26 bits. The extensive experiment shows that network also learns to disentangle original messages and to some extent explains how transmission works . 

\begin{acks}
This work is supported by National Natural Science Foundation of China under Grant U20B2051, 62072114, U20A20178, U1936214.
\end{acks}

\bibliographystyle{ACM-Reference-Format}
\bibliography{ref}

\end{document}